\pdfoutput=1
\documentclass[11pt]{article}

\usepackage{emnlp2021}

\usepackage{times}
\usepackage{latexsym}

\usepackage[T1]{fontenc}
\usepackage[utf8]{inputenc}

\usepackage{microtype}

\usepackage[]{algorithm2e}
\usepackage{algorithmic,float}
\usepackage[utf8]{inputenc} % allow utf-8 input
\usepackage[T1]{fontenc}    % use 8-bit T1 fonts
\usepackage{booktabs}       % professional-quality tables
\usepackage{amsfonts}       % blackboard math symbols
\usepackage{nicefrac}       % compact symbols for 1/2, etc.
\usepackage{microtype}      % microtypography
\usepackage{multirow}
\usepackage{caption}
\usepackage{amsmath,amssymb}
\usepackage{dsfont}			% make hollow 1 work
\usepackage{cases}
\usepackage{color}
\usepackage{tikz}
\usepackage{pgfplots, pgfplotstable}
\usetikzlibrary{shapes,arrows,calc}
\usepackage{caption}
\usepackage{textcomp}
\usepackage{enumitem}
\usepackage{import}
\usepackage{CJKutf8}
\usepackage[toc,page]{appendix}
\usepackage[textsize=large]{todonotes}
\usepackage[normalem]{ulem}
\usepackage[scaled=.8]{beramono}
\usepackage{fancyvrb}
\useunder{\uline}{\ul}{}

\renewcommand{\vec}[1]{{\boldsymbol{\mathbf{#1}}}}

\setlist[itemize]{leftmargin=15pt}
\newcommand\blfootnote[1]{%
  \begingroup
  \renewcommand\thefootnote{}\footnote{#1}%
  \addtocounter{footnote}{-1}%
  \endgroup
}

\title{Levenshtein Training for Word-level Quality Estimation}

\author{Shuoyang Ding$\dagger^*$\quad Marcin Junczys-Dowmunt$\ddagger$\quad Matt Post$\dagger$$\ddagger$\quad Philipp Koehn$\dagger$\\
  $\dagger$ Center for Language and Speech Processing, Johns Hopkins University\quad $\ddagger$ Microsoft \\
  \texttt{\{dings, phi\}@jhu.edu}\quad \texttt{\{marcin.junczysdowmunt, mattpost\}@microsoft.com}
}

\begin{document}
\maketitle
\begin{abstract}
We propose a novel scheme to use the \mbox{Levenshtein} Transformer to perform the task of word-level quality estimation.
A Levenshtein Transformer is a natural fit for this task:
trained to perform decoding in an iterative manner, a Levenshtein Transformer can learn to post-edit without explicit supervision.
To further minimize the mismatch between the translation task and the word-level QE task, we propose a two-stage transfer learning procedure on both augmented data and human post-editing data.
We also propose heuristics to construct reference labels that are compatible with subword-level finetuning and inference.
Results on WMT 2020 QE shared task dataset show that our proposed method has superior data efficiency under the data-constrained setting and competitive performance under the unconstrained setting.
\end{abstract}

% \notepk{Philipp}
% \notemp{Matt}
% \notemjd{Marcin}

\section{Introduction\blfootnote{$^*$ Shuoyang Ding had a part-time affiliation with Microsoft at the time of this work.}}

Quality estimation (QE) is the task of estimating the quality of translation without access to a human-generated reference.
Most recent advances on quality estimation \cite[][\emph{inter alia}]{rei-etal-2020-comet,thompson-post-2020-automatic,ranasinghe-etal-2020-transquest} focus on estimating the quality of translation on either the corpus or segment-level.
However, in practice, the end-users of machine translation (MT) often call for quality signals on more fine-grained level---the level of individual words in a translation.
Such signals are not only useful for more fine-grained triage of translation quality, but also open up the potential for targeted post-processing and faster human post-editing.

\begin{figure}
    % $\hspace{-1cm}
    \scalebox{0.7}{
    \begin{tabular}{cc}
    \begin{tabular}{l} \vspace{-1.7cm}\\ (a)\end{tabular} & \hspace{-0.3cm}\includegraphics[width=0.62\textwidth]{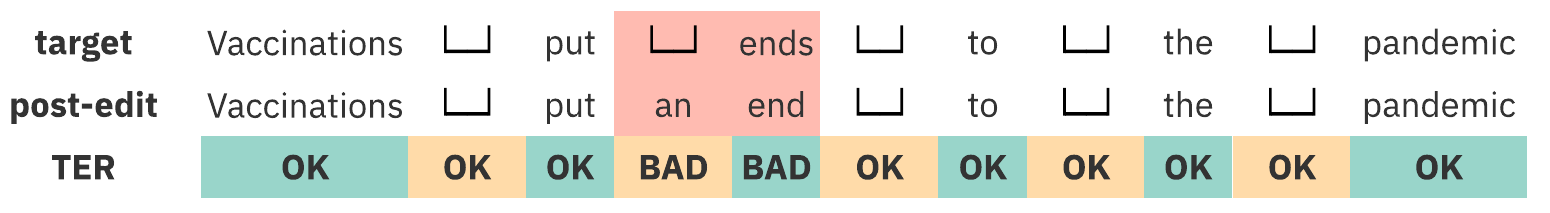} \\
    \begin{tabular}{l} \vspace{-2cm}\\ (b)\end{tabular} & \hspace{-0.6cm}\includegraphics[width=0.63\textwidth]{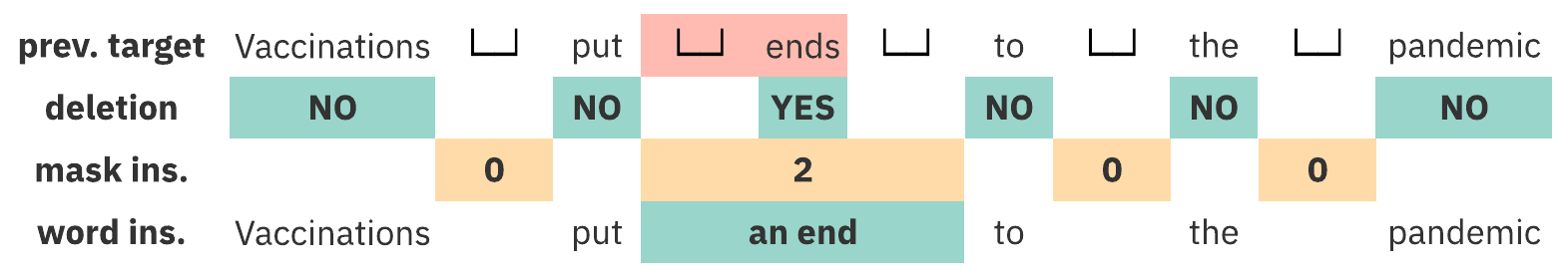} \\
    \end{tabular}
    }
    \caption{Figure (a) shows an example of TER-style edit tags used as reference for word-level quality estimation task. Figure (b) shows a series of hypothetical Levenshtein Transformer edits that generates the same sequence from the target input. The similarity of these edit operations motivates the study in this paper. }
    \label{fig:ter-exp}
\end{figure}

To automatically assess the quality of translations, it is natural to consider starting from a machine translation model, which has already acquired the translation knowledge.
However, it is not clear how to best transfer the translation knowledge to a word-level quality estimation setting, since autoregressive translation models only see the preceding context  of the word they generate.
Hence, they are not well-equipped to perform word-level quality assessments or edits on an existing translation, as there are both preceding and succeeding context for a translated word.
Towards this goal, we leverage Levenshtein Transformer \cite[LevT,][]{DBLP:conf/nips/GuWZ19}, a non-autoregressive neural machine translation (NMT) model trained to generate translations by starting with an empty output sequence and iteratively performing edits on the sequence.
Because of this special training and decoding procedure, the model should have already learned to edit an existing translation sentence without supervision from post-edited translations.
We then use multi-stage transfer learning to teach the model to perform the actual QE task, first on artificially-crafted pseudo post-editing, then on real human post-edited data.

We show that our method achieves better performance than the currently widely adopted Predictor-Estimator scheme \cite{DBLP:journals/talip/KimJKLN17,kepler-etal-2019-openkiwi} under the data-constrained setting, while also being competitive when compared with the high-ranking submissions to the WMT 2020 word-level QE shared task \cite{specia-etal-2020-findings-wmt} under the unconstrained setting.
% We demonstrate that our model compare carefully engineered submissions that are submitted to WMT 2020 word-level QE shared task \cite{specia-etal-2020-findings-wmt}.

% \import{\sectiondir}{levT.tex}
\section{Levenshtein Training for Word-level Quality Estimation}

\subsection{Word-level Quality Estimation}

We follow the problem formulation of word-level quality estimation (QE) in WMT QE shared task \cite{specia-etal-2020-findings-wmt}.
Given the source-side input sentence and MT output pairs,
the participants are asked to perform two binary classification tasks: (1)~whether each word in the target-side translation is correct or not (\emph{translation tag}),
and (2)~whether there are missing words in between each pair of output words (\emph{gap tag}).
The reference tags for such prediction are generated by performing human post-editing on the MT outputs and construct edit alignments with a Translation Error Rate \cite[TER, ][]{snover2006study} computation tool.
An example is shown in Figure \ref{fig:ter-exp}a.
Each submission is evaluated by the Matthews Correlation Coefficient \cite[MCC,][]{matthews1975comparison}.

The state-of-the-art approach to this task is based on the Predictor-Estimator (PredEst) architecture \cite{DBLP:journals/talip/KimJKLN17,kepler-etal-2019-openkiwi}.
At a very high level, the \emph{predictor} training uses a cross-lingual masked language model (MLM) objective, which trains the model to predict a word in the target sentence given the source and both the left and right context on the target side.
An \emph{estimator} is then finetuned from the predictor model to predict word-level QE tags.
In recent years, the top-ranking systems also incorporate large-scale pre-trained crosslingual encoder such as XLM-RoBERTa \cite{conneau-etal-2020-unsupervised}, glass-box features \cite{moura-etal-2020-ist} and pseudo post-editing data augmentation \cite{wei-etal-2020-hw,lee-2020-two}.

\subsection{Levenshtein Transformer}

Intuitively, translation knowledge is very beneficial for the word-level QE task.
Hence, a natural choice for this task is to start from an NMT model and finetune it to produce word-level quality estimation outputs.
However, there are two major limitations of NMT model that makes it unfit for this task: (1)~Most NMT architectures are trained to perform inference in a left-to-right manner, and are therefore ill-equipped to perform edits on an existing translation output; (2)~Most NMT architectures do not have a mechanism to predict whether there words missing at a given location.
% Levenshtein Transformer addresses both of these limitations.
From our introduction below, the readers should notice that Levenshein Transformer successfully addresses both of these limitations.

% These two hurdles led us to choose Levenshtein Transformer as the NMT model to start with.
% Unlike other NMT models, it is exempt from either of these problems: (1) Levenshtein Transformer is trained to perform decoding iteratively; (2) The mask insertion head $\vec{A}_{ins}$ is capable of predicting missing words for word gaps out-of-the-box.
% Hence, we propose to train a Levenshtein Transformer as the initial step towards building a word-level QE model.

The Levenshtein Transformer \cite[LevT,][]{DBLP:conf/nips/GuWZ19} is a neural network architecture that can iteratively generate sequences in a non-autoregressive manner. Unlike normal autoregressive sequence models that have only one prediction head $A_{w}$ to predict the next output words, LevT has two extra prediction heads $A_{del}$ and $A_{ins}$ that predicts deletion and insertion operations based on the output sequence from the previous iteration.
% We now briefly describe the way LevT generates translations, and then the way we use LevT for word-level QE.

For translation generation, at the $k$-th iteration during decoding, with source-side input $\vec{x}$ and target-side sequence input from the previous iteration $\vec{y}^{(k-1)}$ of length $J$, suppose the decoder block output is $\{\vec{h_0}, \vec{h_1}, \ldots, \vec{h_J}\}$.
The following predictions and edits will take place in order:
\begin{itemize}
        \item deletion actions $d_j\in\mathcal{D}^{(k)}$: $$p_{del}^{(k)}(d_j\mid\vec{x}, \vec{y}^{(k-1)}) = \text{softmax}(\vec{A}_{del}^T\,\vec{h_j})$$ for $j\in 1\ldots J_0$ and $J_0 = \left|\vec{y}^{(k-1)}\right|$
        \item mask insertion actions $\gamma_j\in\mathcal{S}^{(k)}$: $$p_{ins}^{(k)}(\gamma_j\mid\vec{x}, \vec{y}') = \text{softmax}(\vec{A}_{ins}^T\,[\vec{h_j}; \vec{h_{j+1}}])$$ for $j\in 0\ldots J_1$, $\vec{y}'=\mathcal{D}^{(k)}(\vec{y}^{(k-1)})$ and $J_1 = \left|\vec{y}'\right|$
        \item word prediction actions $w_j\in\mathcal{W}^{(k)}$: $$p_{w}^{(k)}(w_j\mid\vec{x}, \vec{y}'') = \text{softmax}(\vec{A}_w^T\,\vec{h_j})$$ for $j\in 1\ldots J_2$, $\vec{y}''=\mathcal{S}^{(k)}(\mathcal{D}^{(k)}(\vec{y}^{(k-1)}))$ and $J_2 = \left|\vec{y}''\right|$
\end{itemize}
In the end, $\vec{y}^{(k)} = \mathcal{W}^{(k)}(\mathcal{S}^{(k)}(\mathcal{D}^{(k)}(\vec{y}^{(k-1)})))$.
The iterative process will continue until $\vec{y}^{(k)} = \vec{y}^{(k-1)}$ or a maximum number of iterations is reached.

For the word-level QE task, we only perform one iteration of the above process -- we use the word deletion head $A_{del}$ to predict quality labels for MT words and the mask insertion head $A_{ins}$ to predict quality labels for gaps.
We perform neither the word prediction with $A_w$ nor multiple iterations of prediction.
Still, one should note the similarity between the function of those prediction heads for translation prediction and for word-level QE, which is our motivation for choosing this specific architecture for word-level QE.

It should also be pointed out that using a Levenshtein Transformer translation model as a pre-trained model shares some similar spirit to ELECTRA \cite{DBLP:conf/iclr/ClarkLLM20}, which performs pre-training by learning to detect corrupted tokens generated from a masked language model.

\subsection{Pre-trained Model}\label{sec:m2m-100}
To achieve optimal performance and take advantage of multilingualism, we would also like to take advantage of the large-scale pre-training.
Because there is no pre-trained LevT translation model available, we choose to incorporate M2M-100 \cite{fan2020beyond}, a large-scale pre-trained multilingual autoregressive transformer model.
Recall that the main architectural difference between LevT and a standard transformer model is the two extra prediction heads on LevT.
Hence, to adapt it into a LevT model, we first need to add randomly-initialized extra prediction heads of LevT to the pre-trained checkpoint.
These randomly-initialized prediction heads are then trained with the rest of the model on parallel data in order to adapt the autoregressive translation model to a LevT-style non-autoregressive translation model. 
% We adopt this approach in our experiments and show that initializing with pre-trained models is beneficial to the end-task performance.
% This adapted checkpoint will then be used as the starting point for Levenshtein pre-training to introduce initial knowledge on post-editing.
% \notepk{I found this section very hard to follow. It sounds like you are tweaking M2M which is a very different architecture than non-autoregressive Levenshtein Transformers, by just sprinkling in prediction heads}
% We adopt this approach in our experiments and show that initializing with pre-trained models is beneficial to the end-task performance.

% Despite the popularity of pre-trained MLM-based encoders such as XLM-RoBERTa \cite{conneau-etal-2020-unsupervised}, those models are not trained to generate translations.
% Also, according to previous studies such as \citet{DBLP:conf/emnlp/ClinchantJN19,DBLP:conf/iclr/ZhuXWHQZLL20} have shown, it is not always clear how to best incorporate MLM-based encoders even for regular autoregressive NMT models.

\section{Transfer Learning from Translation to Word-level Quality Estimation}

By training for the translation task, LevT already acquired some initial knowledge of post-editing.
However, there is still some train/inference time mismatch for the word-level QE task regarding (1) the target-side context and (2) the edit tags.
In terms of target-side context, during training, the target-side context is a noisy version of the real target sentence in the training set;
during inference, the target-side context is a translation output from an NMT system.
In terms of edit tags, during training, we want the model to predict LCS edit tags that correct the noisy version of the target sentence;
during inference, we want the model to predict TER-style (Levenshtein) edit tags that correspond to human post-editing of NMT outputs.

To address such mismatch, we adopt a two-step transfer learning scheme.
For both stages of transfer learning, we need \emph{translation triplets} (source, MT output, post-edited output) to perform finetuning.
However, in practice, human post-editing resources are quite scarce. 
Hence, like some previous work \cite{lee-2020-two,wang-etal-2020-hw}, we start by performing transfer learning on synthetic translation triplets, followed by real translation triplets constructed with human post-editing.

\noindent\textbf{Synthetic Data Construction}\quad We explore four different methods for translation triplet synthesis:

\begin{itemize}[leftmargin=*,topsep=0pt] \itemsep -2pt
\item \texttt{src-mt-ref}\quad Take a parallel dataset (\texttt{src, ref}) and translate the source sentence with an MT model (\texttt{mt}).
\item \texttt{bt-rt-tgt}\quad Take a target-side monolingual dataset (\texttt{tgt}). Translate the target sentence with an backward MT model (\texttt{bt}) and then translate \texttt{bt} again with an forward MT model, thus creating a round-trip translated output (\texttt{rt}). Use \texttt{rt} as the MT output and the original \texttt{tgt} as the pseudo post-edited output.
\item \texttt{src-mt1-mt2}\quad Take a source-side monolingual dataset (\texttt{src}). Translate the source sentence with a weaker MT model (\texttt{mt1}) and a stronger MT model (\texttt{mt2}). Use \texttt{mt1} as the MT output and \texttt{mt2} as the pseudo post-edited output.
\item \texttt{mvppe}\quad Take the source-side of a parallel dataset (\texttt{src}). Translate the source sentence with a multilingual MT model as the MT output (\texttt{mt}) and build a pseudo post-edited output (\texttt{pe}) with multiview pseudo post-edit (MVPPE) decoding, as described below.
\end{itemize}

\noindent\textbf{Multiview Pseudo Post-Editing (MVPPE)}\quad Inspired by \citet{thompson-post-2020-paraphrase} which used a multilingual translation system as a zero-shot paraphraser, we propose a novel pseudo post-editing method to build synthetic post-editing dataset from parallel corpus (\texttt{src}, \texttt{tgt}). The first step is to translate the source side of the parallel corpus with a multilingual translation system as the MT output (\texttt{mt}) in the triplet. We then generate the pseudo-post-edited output by ensembling two different \emph{views} of the same model. These two views are:

\begin{itemize}[leftmargin=*,topsep=0pt] \itemsep -2pt
\item the translation output distribution $p_{t}(\text{\tt pe} \mid \text{\tt src})$, with \texttt{src} as the model input;
\item the paraphrase output distribution $p_{p}(\text{\tt pe} \mid \text{\tt tgt})$, with \texttt{tgt} as the model input.
\end{itemize}

Note that both views will create a distribution in the target language space, which can be ensembled in the same way as standard MT model ensembles, forming a interpolated distribution: $$p(\text{\tt pe}\mid \text{\tt src}, \text{\tt tgt}) \triangleq \lambda_t p_t(\text{\tt pe}\mid \text{\tt src}) + \lambda_p p_p(\text{\tt pe}\mid \text{\tt tgt})$$ with $\lambda_t$ and $\lambda_p$ as the interpolation weights.
Similarly, beam search can also be performed on top of the ensemble.
The intuition behind this idea is that such ensemble should create a target sentence that is semantically equivalent to the target side of the parallel corpus, while being close to the original MT output as much as possible, imitating the way humans perform the post-editing task.

\noindent\textbf{Compatibility with Subwords}\quad To the best of our knowledge, previous work on word-level quality estimation either builds models that directly output word-level tags~\cite{lee-2020-two,hu-etal-2020-niutrans-system,moura-etal-2020-ist} or uses simple heuristics to re-assign word-level tags to the first subword token~\cite{wang-etal-2020-hw}.
Since LevT predicts edits on a subword-level starting from translation training, we need to: (1) for inference, convert subword-level tags predicted by the model to word-level tags for evaluation, and (2) for both finetuning stages, build subword-level reference tags.

For inference, the conversion can be easily done by heuristics.
For finetuning, a \emph{naive subword-level tag reference} can be built by running TER alignments on MT and post-edited text after subword tokenization.
However, a preliminary analysis shows that such reference introduces a 10\% error after being converted back to word-level.
Hence, we introduce another heuristic to create \emph{heuristic subword-level tag references}.
The high-level idea is to interpolate the word-level and the naive subword-level references to ensure that the interpolated subword-level tag reference can be perfectly converted back to the word-level references.
Details for the subword tag conversions can be found in Appendix \ref{sec:subword-algos} and Algorithm \ref{algo1} and \ref{algo2} in the appendix.

% \noindent \textbf{Inference}\quad Algorithm \ref{algo1} shows our procedure to convert subword-level tags produced by the model to word-level tags.

% \noindent \textbf{Finetuning}\quad Intuitively, we can build subword-level tag reference by directly using TER alignments built on MT and post-edited text after subword tokenization (BPE, sentencepiece, etc.), which we refer to as naive subword-level reference from now on.
% However, since subword tokenization interacts with the word order error detection (sometimes referred to as ``shifts'') in the TER computation, the naive subword-level references computed this way cannot be perfectly converted back to the word-level references using Algorithm \ref{algo1} introduced above.\footnote{A preliminary analysis shows that the difference is approximately 10\%.}
% Since our model will eventually be evaluated on word-level tags references, finetuning against such naive references is not ideal.

% We propose to build an alternative reference that allows perfect conversion back to their word-level counterparts.
% The high-level idea is to heuristically interpolate the word-level references and the naive subword-level references to ensure the interpolated subword-level tag reference can be perfectly converted back to the word-level references.
% We refer to this interpolated version as \emph{heuristic subword-level reference}.
% The detailed algorithm is shown in Algorithm \ref{algo2}.

% \note{Better name for heuristic subword-level reference?}

\section{Experiments}

\subsection{Setup}

Our experiments are based on the setup of the WMT 2020 QE shared task \cite{specia-etal-2020-findings-wmt}.
The results are reported under two settings: the constrained setting and the unconstrained setting.
Apart from the official metric MCC, we also report the F1 score of the \texttt{OK} and \texttt{BAD} tags (F1-OK and F1-BAD in the result tables).

In the constrained setting, we focus on the data efficiency of our model and use only the human-labeled dataset, the NMT model, and the parallel data (used to train the NMT model) provided by the shared task, with neither large-scale pre-trained model nor synthetic data finetuning.
In the unconstrained setting, we additionally use some extra resources we have access to.
For en-de, we use WMT20 en-de parallel data to train LevT model instead of the smaller parallel data from the shared task, as in the constrained setting.
For en-zh, we use the same dataset because it is already close to the WMT data scale.
We also experiment with the M2M-100-small initialization \cite[][418M parameters]{fan2020beyond} as described in Section \ref{sec:m2m-100}.
Note that M2M-100 directly applies sentencepiece on untokenized data, a tokenization scheme that is incompatible with the shared task setting.
For our experiments with M2M-100-small, we proceed with applying sentencepiece on tokenized data during finetuning.
We also experiment with synthetic data finetuning with different data synthesis methods on en-de language pair, while for en-zh, because we don't have access to an extra high-quality MT system, we only experiment with \texttt{mvppe} method.
Details for our data synthesis setup can be found in Appendix \ref{sec:synthesis}.

\subsection{Results}

\begin{table}[]
\centering
\scalebox{0.9}{
\begin{tabular}{@{}lrrr@{}}
\toprule
              & \textbf{MCC}         & \textbf{F1-OK}       & \textbf{F1-BAD}      \\ \midrule
\textbf{en-de} & \multicolumn{1}{l}{} & \multicolumn{1}{l}{} & \multicolumn{1}{l}{} \\
OpenKiwi       & 0.358                & 0.879                & 0.468                \\
LevT w/o KD    & 0.441                & 0.926                & 0.498                \\
LevT           & \textbf{0.477}       & \textbf{0.929}       & \textbf{0.535}       \\ \midrule
\textbf{en-zh} &                      &                      &                      \\
OpenKiwi       & 0.509                & 0.849                & 0.658                \\
LevT           & \textbf{0.629}       & \textbf{0.885}       & \textbf{0.741}       \\ \bottomrule
\end{tabular}
% \begin{tabular}{@{}lrrr|rrr@{}}
% \toprule
%             & \multicolumn{3}{c|}{\textbf{en-de}}               & \multicolumn{3}{c}{\textbf{en-zh}}                                                                          \\
%             & \textbf{MCC}   & \textbf{F1-OK} & \textbf{F1-BAD} & \multicolumn{1}{l}{\textbf{MCC}} & \multicolumn{1}{l}{\textbf{F1-OK}} & \multicolumn{1}{l}{\textbf{F1-BAD}} \\ \midrule
% OpenKiwi    & 0.358          & 0.879          & 0.468           & 0.509                            & 0.849                              & 0.658                               \\
% LevT w/o KD & 0.441          & 0.926          & 0.498           & \textbf{}                        & \textbf{}                          & \textbf{}                           \\
% LevT        & \textbf{0.477} & \textbf{0.929} & \textbf{0.535}  & \textbf{0.550}                   & \textbf{0.850}                     & \textbf{0.689}                      \\ \bottomrule
% \end{tabular}
}
\caption{Constrained setting. All LevT models here are transformer-base models. F1-OK and F1-BAD are F1 scores of the \texttt{OK} and \texttt{BAD} tags, respectively. Higher is better for all metrics in this table.\label{tab:cons}}
\end{table}

Table \ref{tab:cons} shows results under the data-constrained setting.
Even without knowledge distillation (KD) during LevT training, our model already scores much higher than the baseline OpenKiwi system.
When training with KD data generated with the shared task NMT system (trained on the same parallel data), the advantage of the LevT expands even more.
This shows that our proposal to build a word-level QE system from LevT translation models has higher data efficiency than the widely adopted PredEst approach used by the OpenKiwi baseline.

\begin{table}[]
\centering
\scalebox{0.75}{
\begin{tabular}{@{}lllrrr@{}}
\toprule
\textbf{Init.} & \textbf{LevT} & \textbf{Data Synth.} & \multicolumn{1}{l}{\textbf{MCC}} & \multicolumn{1}{l}{\textbf{F1-OK}} & \multicolumn{1}{l}{\textbf{F1-BAD}} \\ \midrule
\textbf{en-de} &               &                      & \multicolumn{1}{l}{}             & \multicolumn{1}{l}{}               & \multicolumn{1}{l}{}                \\
N              & base          & N                    & 0.539                            & 0.925                              & 0.613                               \\
N              & base          & src-mt-ref           & 0.542                            & 0.925                              & 0.616                               \\
N              & base          & bt-rt-tgt            & 0.535                            & 0.925                              & 0.609                               \\
N              & base          & mvppe                & 0.548                            & 0.926                              & 0.620                               \\
N              & base          & src-mt1-mt2          & 0.549                            & 0.926                              & 0.622                               \\
N              & big           & N                    & 0.551                            & 0.927                              & 0.623                               \\
N              & big           & src-mt1-mt2          & 0.562                            & \textbf{0.939}                     & 0.617                               \\
M2M            & 418M          & N                    & 0.583                            & 0.932                              & 0.650                               \\
M2M            & 418M          & src-mt1-mt2          & \textbf{0.589}                   & 0.934                              & \textbf{0.654}                      \\ \midrule
\textbf{en-zh} &               &                      & \multicolumn{1}{l}{}             & \multicolumn{1}{l}{}               & \multicolumn{1}{l}{}                \\
N              & base          & N                    & 0.629                            & 0.885                              & 0.741                               \\
N              & big           & N                    & 0.625                            & 0.885                              & 0.738                               \\
M2M            & 418M          & N                    & 0.633                            & 0.884                              & 0.744                               \\
M2M            & 418M          & mvppe                & \textbf{0.646}                   & \textbf{0.892}                     & \textbf{0.752}                      \\ \midrule
\textbf{WMT20} &               &                      & \multicolumn{1}{l}{}             & \multicolumn{1}{l}{}               & \multicolumn{1}{l}{}                \\
\multicolumn{3}{l}{\hspace{-0.3cm} en-de best}                        & 0.597                            & 0.935                              & 0.662                               \\
\multicolumn{3}{l}{\hspace{-0.3cm} en-zh best}                        & 0.610                            & 0.887                              & 0.723                               \\ \bottomrule
\end{tabular}
}
\caption{Unconstrained setting. \texttt{base} and \texttt{big} stand for the transformer-base and transformer-big architecture. \texttt{418M} is the M2M-100-small model.\label{tab:uncons}}
\end{table}

Table \ref{tab:uncons} shows results under the unconstrained setting.
We first notice that finetuning with synthetic data before human post-edited data almost always helps.
With the transformer-base model on en-de language pair, we experimented with all four data synthesis methods, and we find that \texttt{src-mt1-mt2} performs the best, closely followed by \texttt{mvppe}.
This might be related to the fact that all LevT models are trained with KD data.
Because of this, the model is better posed to fit synthesized data with MT-like output as pseudo post-edited data, instead of human-generated translations.
Also, initializing with the M2M-100 model is helpful despite the tokenization scheme mismatch, although the performance gain is much more modest on en-zh language pair.
This is possibly influenced by the relatively low translation quality of M2M-100 model on en-zh language pair, as pointed out by \citet{fan2020beyond}.

With all our techniques applied, our best Target MCC result is only slightly behind the winning system on en-de language pair, while being significantly better than the winning system on en-zh language pair.
Most notably, for en-zh language pair, even our smallest LevT model are able to beat the state-of-the-art.
It should also be pointed out that all of our results are achieved without any model ensemble, and our pre-trained model architecture is just a transformer-big counterpart, while other participating teams deployed larger models.

\begin{table}[]
\centering
\scalebox{0.8}{
\begin{tabular}{@{}lrrr@{}}
\toprule
\textbf{Ablation Configuration} & \multicolumn{1}{l}{\textbf{MCC}} & \multicolumn{1}{l}{\textbf{F1-OK}} & \multicolumn{1}{l}{\textbf{F1-BAD}} \\ \midrule
% best                            & 0.583                            & 0.941                              & 0.637                               \\
best                            & 0.589                            & 0.934                              & 0.654                               \\
-LevT                           & 0.555                            & 0.938                              & 0.610                               \\
-LevT +lang-adapt               & 0.565                            & 0.930                              & 0.635                               \\
-LevT -synth.                   & 0.321                            & 0.915                              & 0.380                               \\
-LevT -synth. +lang-adapt       & 0.451                            & 0.946                              & 0.498                               \\ \midrule
-m2m                            & 0.562                            & 0.939                              & 0.617                               \\
-m2m -KD                        & 0.526                            & 0.933                              & 0.589                               \\
-m2m -heuristic tag             & 0.551                            & 0.936                              & 0.610                               \\
-m2m -synth.                    & 0.551                            & 0.927                              & 0.623                               \\
-m2m -synth. -heuristic tag     & 0.539                            & 0.925                              & 0.613                               \\ \bottomrule
\end{tabular}
}
\caption{Ablation analysis. All results trained with synthetic data in this table use the \texttt{src-mt1-mt2} data synthesis method. \texttt{+lang-adapt} stands for adding an extra autoregressive MT training step using the same parallel training data as LevT training, so the M2M-100 model is adapted to translating a specific language pair. \label{tab:ablation}}
\end{table}

To confirm that each component of our training scheme is necessary, we conducted a comprehensive ablation study on en-de language pair, shown in Table \ref{tab:ablation}.
The upper part of the table demonstrates that LevT training is necessary, and we do so by conducting the finetuning directly on M2M-100-small initialization.
Despite the strength of the M2M-100 model as a translation model, there is still a significant performance drop without LevT Training, and more so without synthetic finetuning.
To rule out the effect of bilingual knowledge introduced with LevT training, we also experimented with continue-training the M2M-100 model (\texttt{+lang-adapt} in Table \ref{tab:ablation}) with the same parallel data used for LevT training, but the performance gap remains.
On the other hand, the lower part of the table highlights the effect of various other training techniques, where we use the best system without M2M-100-small initialization as the base.
We can conclude that KD is crucial for optimal performance and that finetuning with heuristic subword-level tag reference is responsible for a small but stable performance improvement.

\section{Conclusion}

In this work, we proposed to use Levenshtein Transformer to substitute the usual MLM-style training in the Predictor-Estimator framework as the initial training step.
We also proposed a series of techniques to effectively transfer the translation knowledge to the word-level QE task, including data synthesis, heuristic subword-level reference, and incorporating pre-trained translation models.
Our results demonstrate superior data efficiency under the data-constrained setting and competitive performance under the unconstrained setting.
We also hope this work can inspire further exploration for other uses of Levenshtein Transformer apart from the non-autoregressive translation.
% \notepk{move statements about performance to the end, focus on novel methods first}

% Future work includes exploring data synthesizing methods that do not require multiple translation systems to operate.
% It is also interesting to deploy the proposed method for other similar sequence editing tasks such as automatic post-editing (APE), grammatical error correction (GEC), and paraphrasing.

\section*{Acknowledgements}
This material is based upon work supported by the United States Air Force under Contract  No. FA8750-19-C-0098. Any opinions, findings, and conclusions or recommendations expressed in this material are those of the author(s) and do not necessarily reflect the views of the United States Air Force and DARPA.

\bibliography{anthology,custom}
\bibliographystyle{acl_natbib}

\appendix
\clearpage
% \section{Details on Levenshtein Transformer Inference \label{sec:levT-details}}
% For the $k$-th iteration during decoding, with source-side input $\vec{x}$ and target-side sequence input $\vec{y}^{(k-1)}$ of length $J$, suppose the decoder block output is $\{\vec{h_0}, \vec{h_1}, \ldots, \vec{h_J}\}$.
% The following predictions and edits will take place in order:
% \begin{itemize}
% 	\item deletion actions $d_j\in\mathcal{D}^{(k)}$: $$p_{del}^{(k)}(d_j\mid\vec{x}, \vec{y}^{(k-1)}) = \text{softmax}(\vec{A}_{del}^T\,\vec{h_j})$$ for $j\in 1\ldots J_0$ and $J_0 = \left|\vec{y}^{(k-1)}\right|$
% 	\item mask insertion actions $\gamma_j\in\mathcal{S}^{(k)}$: $$p_{ins}^{(k)}(\gamma_j\mid\vec{x}, \vec{y}') = \text{softmax}(\vec{A}_{ins}^T\,[\vec{h_j}; \vec{h_{j+1}}])$$ for $j\in 0\ldots J_1$, $\vec{y}'=\mathcal{D}^{(k)}(\vec{y}^{(k-1)})$ and $J_1 = \left|\vec{y}'\right|$
% 	\item word prediction actions $w_j\in\mathcal{W}^{(k)}$: $$p_{w}^{(k)}(w_j\mid\vec{x}, \vec{y}'') = \text{softmax}(\vec{A}_w^T\,\vec{h_j})$$ for $j\in 1\ldots J_2$, $\vec{y}''=\mathcal{S}^{(k)}(\mathcal{D}^{(k)}(\vec{y}^{(k-1)}))$ and $J_2 = \left|\vec{y}''\right|$
% \end{itemize}
% In the end, $\vec{y}^{(k)} = \mathcal{W}^{(k)}(\mathcal{S}^{(k)}(\mathcal{D}^{(k)}(\vec{y}^{(k-1)})))$. The iterative process will continue until $\vec{y}^{(k)} = \vec{y}^{(k-1)}$ or a maximum number of iterations is reached.

\section{Details on Edit Tag Conversion to/from Subword-Level \label{sec:subword-algos}}

Algorithm \ref{algo1} shows the tag conversion algorithm from subword-level to word-level (for inference). Algorithm \ref{algo2} shows the tag conversion algorithm from word-level to subword-level (for finetuning).

\RestyleAlgo{ruled}
\begin{algorithm}[h]
 \KwIn{subword-level token sequence $\vec{y}^{sw}$, word-level token sequence $\vec{y}^w$, subword-level tag sequence $\vec{q}^{sw}$}
 \KwOut{word-level tag sequence $\vec{q}^w$}
 $\vec{q}^w \leftarrow$ []\;
 \For{each word $w_k$ in $\vec{y}^w$}{
    find subword index span $(s_k, e_k)$ in $\vec{y}^{sw}$ that corresponds to $w_k$\;
    $\vec{q}^{sw}_{k} \leftarrow$ subword-level translation and gap tags within span $(s_k, e_k)$\;
    $g^{sw}_{s_k} \leftarrow$ subword-level gap tag before span $(s_k, e_k)$\tcp*{$g^{sw}_{s_k-1}\in\vec{q}^{sw}$}
    \eIf{$\forall\,\vec{q}^{sw}_k$ are \texttt{OK}}{
        $\vec{q}^w$ += [$g^{sw}_{s_k}$, \texttt{OK}]\;
    }{
        $\vec{q}^w$ += [$g^{sw}_{s_k}$, \texttt{BAD}]\;
    }
 }
 $\vec{q}^w$ += [$\vec{q}^{sw}$[-1]]\tcp*{add ending gap tag}
 \Return{$\vec{q}^w$}\;
 \caption{Conversion of subword-level tags to word-level tags\label{algo1}}
\end{algorithm}

\RestyleAlgo{ruled}
\begin{algorithm}[h]
 \KwIn{subword-level token sequence $\vec{y}^{sw}$, word-level token sequence $\vec{y}^w$, naive subword-level tag sequence $\vec{q}^{sw}$, word-level tag sequence $\vec{q}^{w}$}
 \KwOut{heuristic subword-level tag sequence $\widetilde{\vec{q}^{sw}}$}
 $\widetilde{\vec{q}^{sw}} \leftarrow$ []\;
 \For{each word $w_k$ in $\vec{y}^w$}{
   find subword index span $(s_k, e_k)$ in $\vec{y}^{sw}$ that corresponds to $w_k$\;
   $\vec{q}^{sw}_k \leftarrow$ subword-level translation and gap tags within span $(s_k, e_k)$\;
   $t^w_k \leftarrow$ word-level translation tag for $w_k$\tcp*{$t^w_k\in\vec{q}^w$}
   $g^w_k \leftarrow$ word-level gap tag before $w_k$\tcp*{$g^w_k\in\vec{q}^w$}
   $\widetilde{\vec{q}^{sw}}$ += [$g^w_k$]\tcp*{copy left gap tag}
   $n$ = $\left|\vec{q}^{sw}_k\right|$\tcp*{\# subwords for $w_k$}
   \uIf{$t^t_k$ is \texttt{OK}}{
     \tcc{word is \texttt{OK}, so all subwords and gaps between them should be \texttt{OK}}
     $\widetilde{\vec{q}^{sw}}$ += [\texttt{OK}] * $(2n-1)$
   }
   \uElseIf{$\exists\,$\texttt{BAD} in $\vec{q}^{sw}_k$}{
     \tcc{no conflict between subword-level tag and word-level tag -- copy $\vec{q}^{sw}_k$ as-is}
     $\widetilde{\vec{q}^{sw}}$ += $\vec{q}^{sw}_k$\;
   }
   \uElse{
     \tcc{subword-level tag disagrees with word-level tag -- force it as all-\texttt{BAD} to guarantee perfect conversion}
     $\widetilde{\vec{q}^{sw}}$ += [\texttt{BAD}] * $(2n-1)$
   }
 }
 $\widetilde{\vec{q}^{sw}}$ += [$\vec{q}^{w}$[-1]]\tcp*{add ending gap tag}
 \Return{$\widetilde{\vec{q}^{sw}}$}\;
 \caption{Construction of heuristic subword-level tags\label{algo2}}
\end{algorithm}

\section{Details of Experimental Setup}
\subsection{Data Synthesis\label{sec:synthesis}}
For all en-de experiments, we took the first 1 million line from the Europarl corpus and conduct data synthesis.
\begin{itemize}[leftmargin=*,topsep=3pt] \itemsep -2pt
        \item \texttt{src-mt-ref}\quad We translate the source side of the parallel data with the NMT system provided by the shared task organizer.
	\item \texttt{bt-rt-tgt}\quad Both the back-translation and the round-trip translation are performed with M2M-100-mid (1.2B) model.
        \item \texttt{src-mt1-mt2}\quad We take the source side of the parallel data and translate it with the NMT system from the shared task (weaker system, \texttt{mt1}) and the Facebook winning system for the WMT19 en-de news translation \cite[][stronger system, \texttt{mt2}]{ng-etal-2019-facebook}. We remove all the cases where \texttt{mt1} and \texttt{mt2} are identical. 
	\item \texttt{mvppe}\quad The MVPPE decoding is conducted with M2M-100-mid (1.2B) model.
\end{itemize}

For en-zh experiments, we take the shared task en-zh parallel data but exclude the UN data for MVPPE data synthesis. The same multilingual translation model is used.

We also experimented with using larger synthetic data for en-de with some synthesis method, but didn't observe a significant performance difference compared to this smaller dataset.

% \note{Another interesting but short analysis is that you don't need optimal BLEU to get best word-level QE performance}

\subsection{Misc.}
We preprocess our data by first tokenizing with Moses tokenizer, and then applying subword segmentation.
For all the LevT models without M2M-100 initialization, we use the same BPE model and source/target-side vocabulary as the official NMT checkpoint provided by the WMT20 QE shared task.
For models with M2M-100 initialization, we use the M2M-100 sentencepiece model.

Under the constrained setting, we use the NMT checkpoint supplied by the shared task to generate the knowledge distillation data for LevT translation training, both for en-de and en-zh.
Under the unconstrained setting, for en-de, we use the Facebook winning system for the WMT19 en-de news translation, and for en-zh, we use our own Transformer-base en-zh model trained on WMT17 en-zh data.

All of our implementations are based on the Fairseq toolkit.
We use the same hyperparameter for LevT translation model training as the document provided in Fairseq\footnote{\url{https://github.com/pytorch/fairseq/blob/master/examples/nonautoregressive_translation/README.md}}.
For both synthetic and human post-edited data finetuning, we use Adam optimizer with a learning rate 2e-5 with warmup (4000 updates for synthetic finetuning and 2000 for human post-edited data finetuning), and we use the shared task development set to select the best checkpoint.

For all the \texttt{mvppe} experiments, we use $\lambda_t = 2.0$ and $\lambda_p = 1.0$, after doing a grid search over $\lambda_t=\{1.0, 2.0, 3.0\}$ and $\lambda_p = \{1.0, 1.2, 1.5\}$ with a goal to match the TER distribution of human post-editing obtained from the en-de human PE dev data.

All the word-level and subword-level tags we use as the reference for finetuning are computed using our own TER implementation\footnote{\url{https://github.com/marian-nmt/moses-scorers}}, but we stick to the original reference tags in the test set for evaluation to avoid potential result mismatch. Our evaluations are done with the official evaluation scripts\footnote{\url{https://github.com/sheffieldnlp/qe-eval-scripts}} from the shared task. The script computes the Matthews Correlation Coefficient \cite[MCC,][]{matthews1975comparison}, which is formulated as follows:
\begin{align}
    S &= \dfrac{TP + FN}{N} \\
    P &= \dfrac{TP + FP}{N} \\
    MCC &= \dfrac{TP / N - S\times P}{\sqrt{PS(1 - S)(1 - P)}}
\end{align}
where $TP$/$FP$ stands for true/false positives and $TN$/$FN$ stands for true/false negatives.
$N$ stands for the number of examples in the dataset.
The script also computes F1 scores of the \texttt{OK} and \texttt{BAD} tags.

Table \ref{tab:data} shows some statistics of the data we use in our experiments.

\begin{table}[]
\scalebox{0.8}{
\begin{tabular}{@{}lr@{}}
\toprule
                                 & \textbf{\# sentence pairs/triplets} \\ \midrule
shared task en-de parallel       & 3.96M                              \\
shared task en-zh parallel       & 20.3M                              \\
WMT20 en-de parallel             & 44.2M                              \\
en-de \texttt{src-mt-ref} synthetic    & 945K                         \\
en-de \texttt{src-mt1-mt2} synthetic   & 808K                         \\
en-de \texttt{bt-rt-tgt} synthetic     & 998K                         \\
en-de \texttt{mvppe} synthetic         & 993K                         \\
en-zh \texttt{mvppe} synthetic   & 9.26M			      \\
shared task en-de human PE train & 7K                                 \\
shared task en-zh human PE train & 7K                                 \\
shared task en-de human PE dev   & 1K                                 \\
shared task en-zh human PE dev   & 1K                                 \\
shared task en-de human PE test  & 1K                                 \\
shared task en-zh human PE test  & 1K                                 \\ \bottomrule
\end{tabular}
}
\caption{Data Statistics for Our Experiments \label{tab:data}}
\end{table}

\subsection{Extra Results on the Updated Dataset}

As of Sep. 2021, there is an updated version of the WMT20 shared task dataset (train/dev/test) with the same MT output but different human post-edited output.
For reference of future work, we provide results on this updated dataset from some of our experiment configurations in Table \ref{tab:updated-ds}.

\begin{table}[]
\centering
\scalebox{0.9}{
\begin{tabular}{@{}lrrr@{}}
\toprule
                    & \textbf{MCC}         & \textbf{F1-OK}       & \textbf{F1-BAD}      \\ \midrule
\textbf{en-de}      & \multicolumn{1}{l}{} & \multicolumn{1}{l}{} & \multicolumn{1}{l}{} \\
LevT base (constr.) & 0.363                & 0.951                & 0.410                \\
LevT base           & 0.434                & 0.954                & 0.479                \\
LevT big            & 0.459                & \textbf{0.958}       & 0.498                \\
LevT m2m            & 0.489                & 0.955                & 0.533                \\
LevT m2m + synth.   & \textbf{0.500}       & 0.956                & \textbf{0.544}       \\
\textbf{en-zh}      &                      &                      &                      \\
LevT base           & \textbf{0.463}       & \textbf{0.907}       & \textbf{0.554}       \\
LevT big            & 0.460                & 0.903                & 0.552                \\
LevT m2m            & 0.453                & 0.898                & 0.548                \\
LevT m2m + synth.   & 0.459                & 0.900                & 0.552                \\ \bottomrule
\end{tabular}
}
\caption{Results on the updated dataset \label{tab:updated-ds}}
\end{table}

\end{document}